\documentclass[12pt]{article}

\usepackage{newtxtext,newtxmath}

\usepackage{graphicx}

\usepackage[letterpaper,margin=0.8in]{geometry}

\renewenvironment{abstract}
	{\quotation}
	{\endquotation}

\date{}

\makeatletter
\renewcommand{\fnum@figure}{\textbf{Figure \thefigure}}
\renewcommand{\fnum@table}{\textbf{Table \thetable}}
\makeatother

\usepackage{scicite}

\usepackage{url}

\def\scititle{Geometry-based pneumatic actuators for soft robotics
}
\title{\bfseries \boldmath \scititle}

\author{
	Rui~Chen$^{1\ast}$,
	Daniele~Leonardis$^{1}$,
	Domenico~Chiaradia$^{1}$,
	Antonio~Frisoli$^{1}$\and
	\small$^{1}$Institute of Mechanical Intelligence, School of Advanced Studies Sant'Anna (SSSA), 56127 Pisa, Italy.\and
	\small$^\ast$Corresponding author. Email: rui.chen@santannapisa.it
}

\begin{document} 

\maketitle

\begin{abstract} \bfseries \boldmath

Soft pneumatic actuators enable safe human-machine interaction with lightweight and powerful applied parts. On the other side, they suffer design limitations as regards complex actuation patterns, including minimum bending radii, multi-states capabilities and structural stability. We present geometry-based pneumatic actuators (GPAs), a design and implementation approach that introduces constraint layers with configurable CNC heat-sealed chambers. The approach achieves predictable deformation, near-zero bending radii, multi-states actuation, and enables customizable and repeatable complex actuated geometries. Mathematical modeling reveals predictable linear angle transformations and validates nonlinear torque-angle relationships across diverse configurations. We demonstrate versatility of the GPAs approach through three applications: a 49-g wrist exoskeleton reducing muscle activity by up to 51\%, a 30.8-g haptic interface delivering 8-N force feedback with fast response, and a 208-g bipedal robot achieving multi-gait locomotion. GPAs establish a configurable platform for next-generation wearable robotics, haptic systems, and soft locomotion devices.

\end{abstract}

\section{Introduction}

Soft robotics has revolutionized human-machine interaction by introducing compliant, adaptive systems that operate safely alongside humans in complex environments\cite{el2020_Soft_Actuator_Review}. These technologies have enabled transformative applications spanning wearable rehabilitation devices\cite{zhou2024_portable_exoskeleton_app,Rui2025_Glove_app}, medical interventions\cite{ze2022S_pinning_Origami_locomotion_medical_app,choi2022_Body_Integreted_medical_app}, haptic interfaces\cite{guo2024_DEA_haptic_APP, du2024_haptiknit_Haptic_app, ha2025_full_freedom_Haptic_app,zhang_2023_Origami_Active_Haptic_app}, environmental exploration\cite{katzschmann2018_hydrolic_Underwater_sea_app,pan2025_locomotion_sea_app}, and autonomous locomotion systems\cite{zhakypov2019_Multi_Locomotion_locomotion_app, shepherd2011_pneumatic_multigait_Locomotion_app,laschi2012_Soft_Octopus_app, fu2025_origami_manipulation_app, wu2021_sStretchable_Origami_arm_app}. This progress has been powered by diverse actuation mechanisms, including dielectric elastomers\cite{guo2024_DEA_haptic_APP, youn2025_DEA_Haptic_app}, combustion systems\cite{aubin2023_combustion_actuators}, phase-change materials\cite{fonseca2025Electrically_phase_transition_actuators}, electromagnetic systems\cite{mao2020_Soft_Electromagnetic_actuator, zhang2025D_Printed_Magnetoactive_Origami_Actuator}, liquid crystal elastomers\cite{yang2025_liquid_Crystal_Elastomer_Fiber_Actuators, li2021_Liquid_Crystal_Elastomers_muscles_actuator}, and hydraulic/pneumatic systems\cite{gravert2024_electrohydraulic_actuators, katzschmann2018_hydrolic_Underwater_sea_app, Rui2025_LPPAMs_actuator_app, tang2020_pneumatic_bistable_Spine_locomotion_app, feng2023_X-PAMs_actuator, niiyama2015_Pouch_Motor}. Among these, pneumatic fabric-based actuators have emerged as particularly promising due to their exceptional combination of lightweight design, inherent compliance, and cost-effective manufacturing\cite{xavier2022Pneuamtic_review}.

Existing fabric-based pneumatic actuators follow two principal design paradigms. The first employs an outer fabric layer that strategically constrains an independent inner bladder, where tailored fabric patterning achieved through sewing\cite{zhu2020_Sewing_fabrication_tube, phan2022Fabrication_sewing_textile, guo2024_Encoded_Sewing_actuator} or knitting\cite{luo2022Fabrication_Machine_knitting, sanchez2023_3D_knitting, cappello2018FabricationAnisotropy,Li_2024_knitting} guides deformation upon inflation. While these techniques enable complex geometries and programmable actuation through engineered pleated structures or asymmetric knitting patterns, they are fundamentally constrained by large minimum bending radii and limited capacity for multi-states actuation. The second paradigm integrates a polymeric coating—typically polyurethane—directly onto fabric substrates to create airtight composites, with pneumatic chambers formed through heat-sealing multiple layers. Manufacturing approaches including laser cutting\cite{amiri2018laser_cut, yamaoka2018_accordionfab_Laser_cut, Wu2025pouch_stack}, pleated architectures\cite{zhang2023_Soft_Fabric_actuator, nishioka2017Fabric_pleated}, and pouch motor designs\cite{suulker2022soft_pouch_motor,rajappan2022_logic_pouch_motor, o2022unfolding_cylinder,yang2020_design_chamber_cylinder} have demonstrated various actuation patterns. However, these implementations face persistent challenges: pleated structures remain limited to large bending radii with constrained programmability, while pouch motors—though capable of tighter bending—result in bulkier configurations with reduced workspace and minimal multi-states capability.

Computer numerical control (CNC) heat sealing represents a transformative fabrication methodology\cite{niiyama2015_Pouch_Motor,gohlke2023wireshape_CNC, goshtasbi2025weld_CNC, rus2015_Design_Fabrication_review}, accessible even with modified laboratory-scale 3D printing hardware. This technique fundamentally shifts the design paradigm from chamber-centric to geometry-centric approaches, enabling arbitrary distribution of bending regions including areas with minimal radii. Yet conventional single-chamber CNC heat-sealed actuators implementing complex or concave geometries suffer from a critical failure mode: catastrophic deformation instabilities under external loading, caused by seal edges propagating into the inflated structure (Fig.~\ref{Fig-M-Illustration}A). This structural vulnerability, combined with the inherent large bending radii of traditional fabric actuators\cite{zhang2023_Soft_Fabric_actuator, sanchez2023_3D_knitting, guo2024_Encoded_Sewing_actuator, phan2022Fabrication_sewing_textile}, severely restricts their deployment in applications where precise, stable actuation is essential.

In this work we introduce geometry-based pneumatic actuators (GPAs)—a design framework that addresses these fundamental limitations through strategic integration of multiple CNC heat-sealed chambers with constraint layers to achieve predetermined geometries upon inflation (Fig.~\ref{Fig-M-Illustration}B). This architecture eliminates deformation instabilities under external loading (Fig.~\ref{Fig-M-Illustration}C and Movie S1), while enabling near-zero bending radii with sharp angular configurations that can closely emulate biological joint mechanics. Beyond structural stability, GPAs provide flat or concave contact surfaces, in contrast to the convex profiles of single-chamber systems, offering critical advantages for ergonomic wearable interfaces\cite{Rui2025_Glove_app}. The strategic design and combination of chambers with constraint layers unlocks multi-states actuation and multi-degree-of-freedom control, transcending the functional limitations of conventional single-state pneumatic actuators. 

We present here the mathematical modeling of GPAs behavior, comprehensive characterization of implemented prototypes, and validation across three paradigmatic applications: exoskeletons, haptic interfaces, and autonomous locomotion systems (Fig.~\ref{Fig-M-Illustration}D). These key features—deformation stability, multi-states capability, minimal bending radii, and customizable geometry—establish GPAs as a transformative platform for next-generation soft robotic systems.

\section*{Results}

\subsection*{Design principles of geometry-based pneumatic actuators}

\begin{figure}[htbp]
    \centering
    \includegraphics[width=0.95\textwidth]{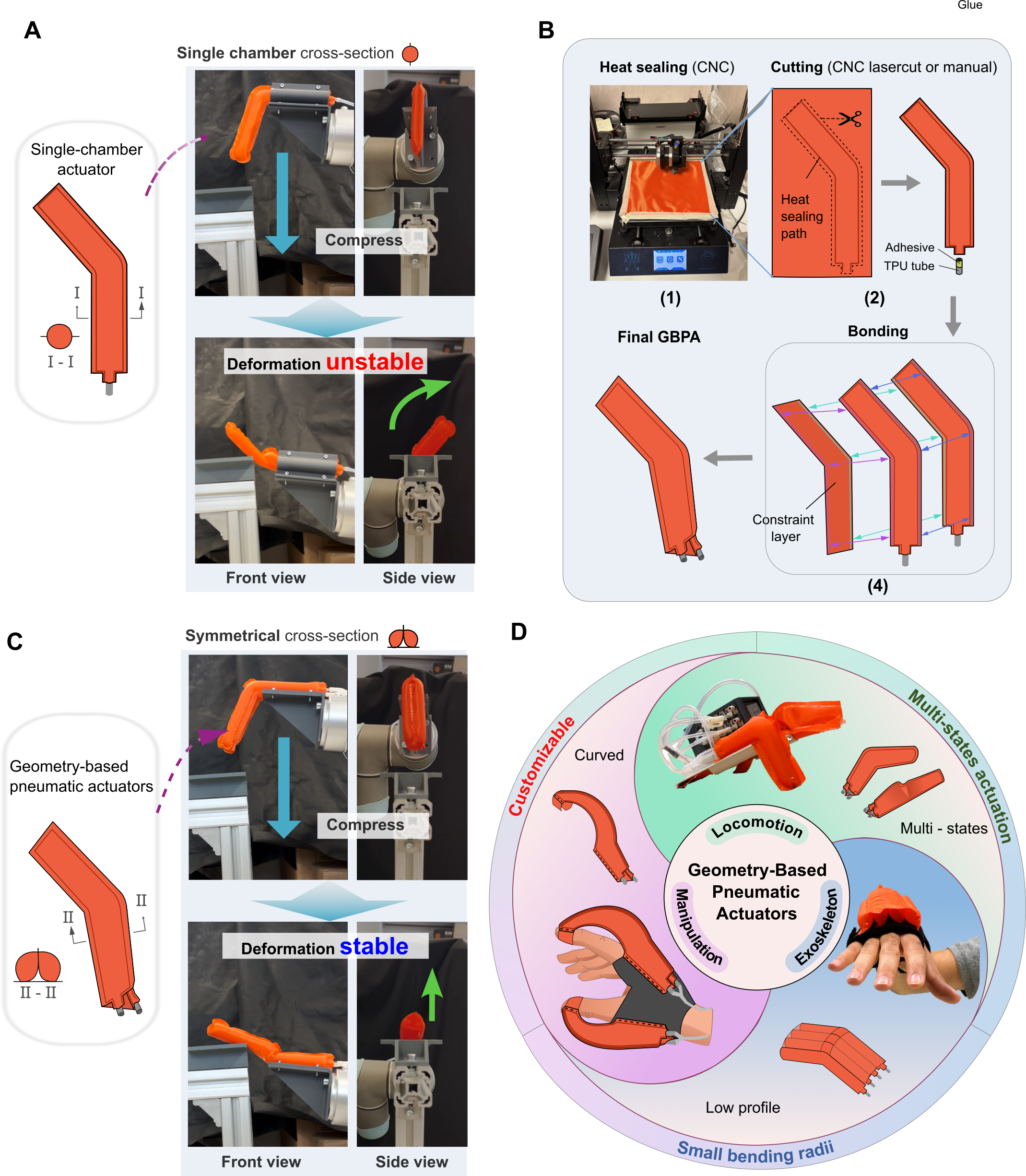}
    \caption{\textbf{Geometry-based pneumatic actuators overcome fundamental limitations of conventional designs.}
    \textbf{(A)} Single-chamber actuators exhibit unpredictable deformation instabilities under external loading. \textbf{(B)} GPAs architecture integrates constraint layers with CNC heat-sealed chambers for programmable geometry control. \textbf{(C)} Constraint layer integration eliminates deformation instabilities, ensuring predictable actuation behavior. \textbf{(D)} GPAs versatility enables diverse applications: exoskeletons, haptic interfaces, and autonomous locomotion systems.}
    \label{Fig-M-Illustration}
\end{figure}

Conventional CNC heat-sealed single-chamber actuators suffer from a critical failure mode: unpredictable deformation instabilities under external loading that manifest as sudden behavioral transitions driven by unbalanced contact mechanics (Fig.~\ref{Fig-M-Illustration}A). We address this fundamental limitation through the GPAs design framework, which integrates constraint layers that guide deformation along predetermined pathways and eliminate catastrophic failure modes.

The GPAs architecture strategically combines multiple CNC heat-sealed chambers with constraint layers—either partial or complete—attached to the actuator's inner or outer surfaces (Fig.~\ref{Fig-M-Illustration}B). These constraint layers function as geometric guides that direct pneumatic expansion toward specific angular configurations while suppressing the deformation instabilities that plague conventional designs. Comparative testing under identical external loading conditions revealed that single-chamber actuators underwent substantial lateral deviation, whereas GPAs maintained stable behavior entirely within the intended actuation plane (Fig.~\ref{Fig-M-Illustration}C).

This design framework enables rapid customization of actuator geometry through standard CNC heat-sealing processes while unlocking multi-states actuation capabilities previously unattainable with conventional approaches. The resulting GPAs achieve critical performance metrics that address fundamental limitations: stable and predictable deformation behavior under external loading, minimal bending radii enabling compact form factors, multi-states actuation for versatile functionality, and customizable geometries including ergonomic concave interfaces. These features collectively position GPAs as a transformative platform for diverse soft robotic applications spanning multiple operational domains (Fig.~\ref{Fig-M-Illustration}D).

\subsection*{Characterization and analytical modeling of GPAs}

\begin{figure}[htbp]
    \centering
    \includegraphics[width=0.9\textwidth]{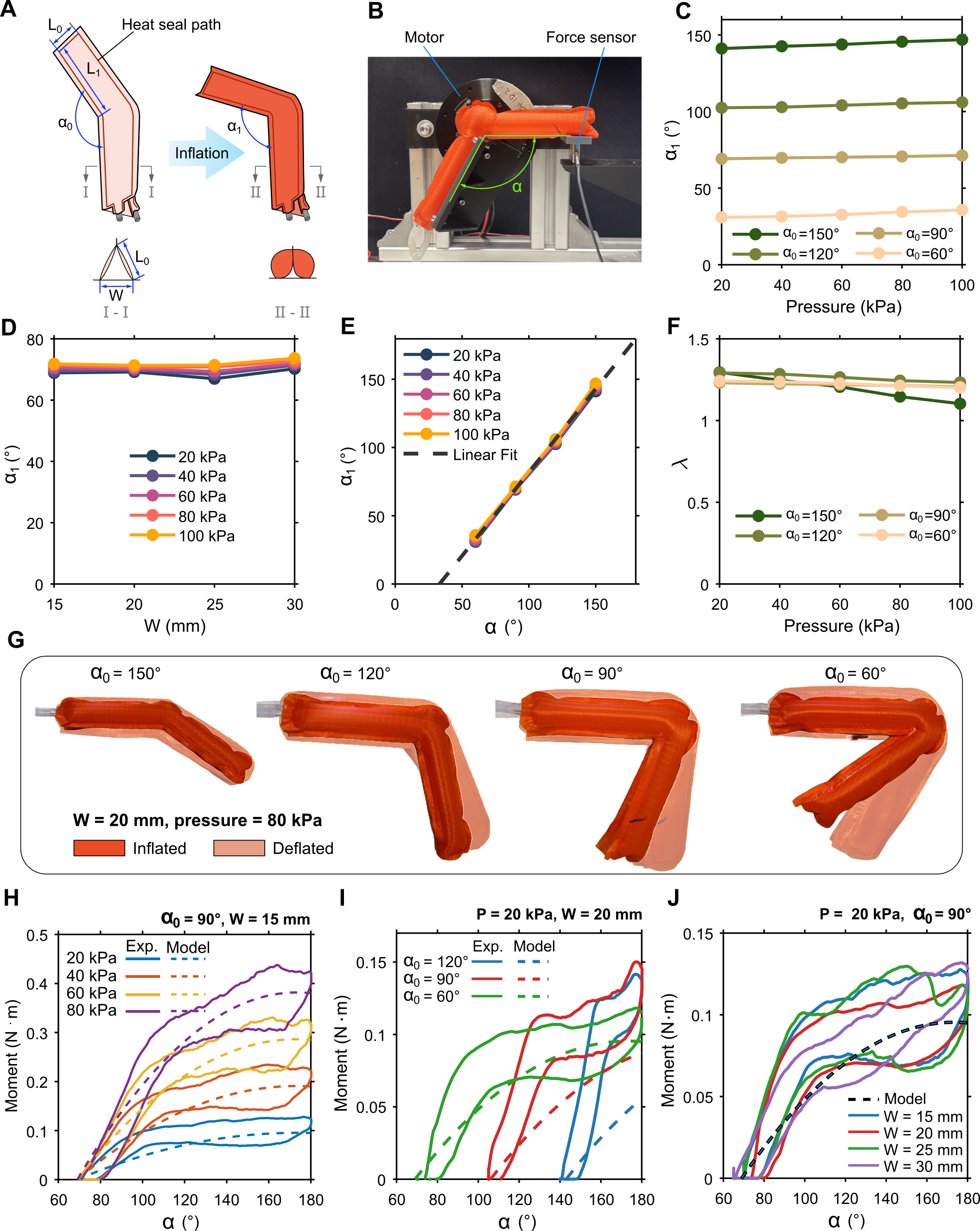}
    \caption{\textbf{Systematic characterization reveals predictable GPAs performance relationships.} \textbf{(A)} Key geometric parameters defining GPAs behavior. \textbf{(B)} Torque characterization system for force-angle relationship determination. \textbf{(C)} Inflation pressure minimally affects final bending angle. \textbf{(D)--(F)} Parametric analysis demonstrating linear relationships between initial angle $\alpha_0$, final angle $\alpha_1$, and contraction factor $\lambda$. \textbf{(G)} Inflation of actuators with different initial angles $\alpha$. \textbf{(H)} Mathematical model validation of nonlinear torque-angle relationships with pressure dependence. \textbf{(I) and (J)} Model validation across diverse actuator configurations.}
    \label{Fig-M-Characterzation}
\end{figure}

We systematically characterized GPAs performance across key geometric and loading parameters to establish predictive design relationships (Fig.~\ref{Fig-M-Characterzation}A). Using dedicated experimental setups for angular measurements (Fig.~S2B) and torque analysis (Fig.~\ref{Fig-M-Characterzation}B), we revealed well-defined relationships between design parameters and actuator behavior.

Parametric studies demonstrated that inflation pressure exerted minimal influence on final bending angle across the operational range (Fig.~\ref{Fig-M-Characterzation}C), while actuator width and circumferential length exhibited negligible effects on angular configuration (Fig.~\ref{Fig-M-Characterzation}D and Supplementary Figs.~S2--S4). Most notably, we identified a robust linear relationship between initial angle $\alpha_0$ and final inflated angle $\alpha_1$, which remained consistent across both GPAs and single-chamber configurations (Fig.~\ref{Fig-M-Characterzation}F and Supplementary Figs.~S2--S3)\cite{Andrade2025_Star_griper}.

We introduced a contraction factor $\lambda = \frac{\pi - \alpha_0}{\pi - \alpha_1}$ (where $0 < \alpha_0 < \pi$) that characterizes the geometric transformation during actuation. This contraction factor exhibited linear dependence on initial angle (Fig.~\ref{Fig-M-Characterzation}E and G) with modest pressure sensitivity (Fig.~\ref{Fig-M-Characterzation}F), enabling predictive design of target geometries through straightforward geometric relationships.

Building on these empirical observations, we developed a simplified mathematical model to predict actuator moment behavior and validated predictions against experimental measurements. Moment characterization revealed nonlinear force-angle relationships, with output moment increasing substantially as bending angle increased (Fig.~\ref{Fig-M-Characterzation}H). Higher pressures generated nearly proportional increases in torque output at equivalent angular positions. The initial angle controlled the final inflated configuration and therefore determined overall output moment capacity (Fig.~\ref{Fig-M-Characterzation}I), while actuator width had negligible effect on moment generation (Fig.~\ref{Fig-M-Characterzation}J). The mathematical model demonstrated excellent agreement with experimental observations across varied parameter combinations, validating our predictive design framework and enabling rational optimization of GPAs for specific application requirements (Fig.~S5).

\subsection*{Extended GPAs architectures for enhanced functionality}

\begin{figure}[htbp]

    \centering
    \includegraphics[width=1\textwidth]{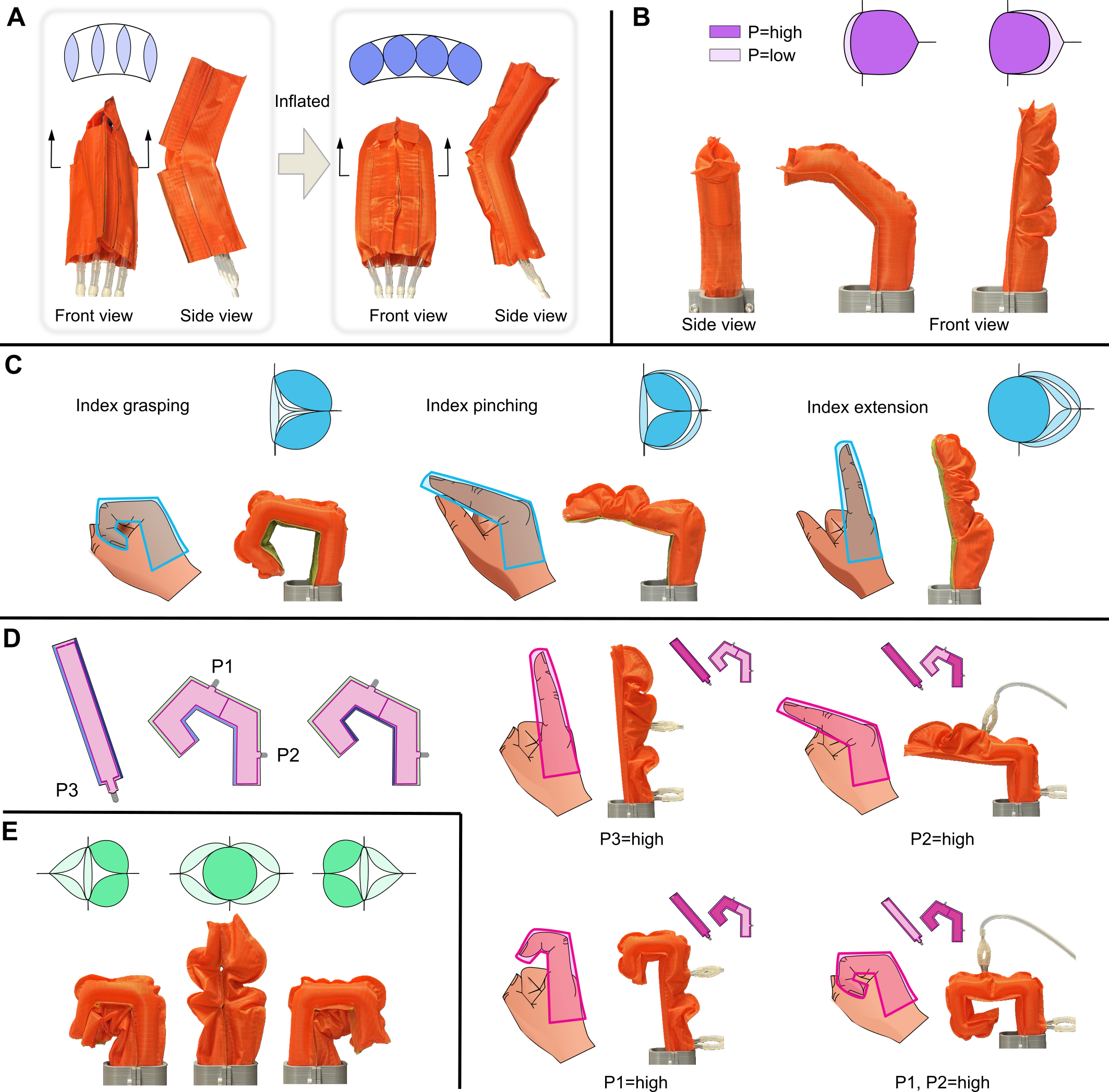}
    \caption{\textbf{Extended GPAs architectures enable specialized functionality.} \textbf{(A)} Parallel configurations achieve compact, low-profile designs. \textbf{(B)} Single-chamber variants demonstrate constraint layer versatility. \textbf{(C)} multi-states architectures enable complex biomimetic motions. \textbf{(D)} Segmented designs provide independent multi-region control. \textbf{(E)} Bilateral configurations permit bidirectional actuation capabilities.}
    \label{Fig-M-Extended}
\end{figure}

The modular nature of the GPAs framework enables sophisticated architectural variations tailored for specialized applications (Movie S2). Parallel configurations constrain multiple CNC heat-sealed chambers with identical geometry in parallel arrangement (Fig.~\ref{Fig-M-Extended}A), achieving compact, low-profile designs ideal for wearable systems requiring lightweight construction and minimal encumbrance. Single-chamber configurations (Fig.~\ref{Fig-M-Extended}B) strategically convert the bottom constraint layer into part of the chambers, enabling shared constraint layers among multiple chambers for enhanced multi-functionality. This architecture enables more efficient volume utilization through strategic chamber integration.

We extended the multiple-chamber concept into multi-states GPAs that incorporate overlapping layer chambers, each responsible for distinct geometric configurations (Fig.~\ref{Fig-M-Extended}C). When configured as finger-like structures, selective inflation of different chambers replicated complex biomimetic motions including extension, pinching, and grasping—enabling advanced manipulation capabilities. multi-states actuation can alternatively be achieved by segmenting a single layer chamber into multiple independently controllable regions for separated joint actuation (Fig.~\ref{Fig-M-Extended}D). These segmented configurations, with additional degrees of freedom under independent control, demonstrated sophisticated gesture replication including pointing, hooking, and nuanced grasping behaviors. Bilateral configurations, achieved by arranging actuators in opposing orientations (Fig.~\ref{Fig-M-Extended}E), enabled bidirectional actuation and expanded workspace capabilities for applications requiring multi-directional motion control.

\subsection*{Lightweight soft exoskeleton for wrist motor assistance}

\begin{figure}[htbp]
    \centering
    \includegraphics[width=1\textwidth]{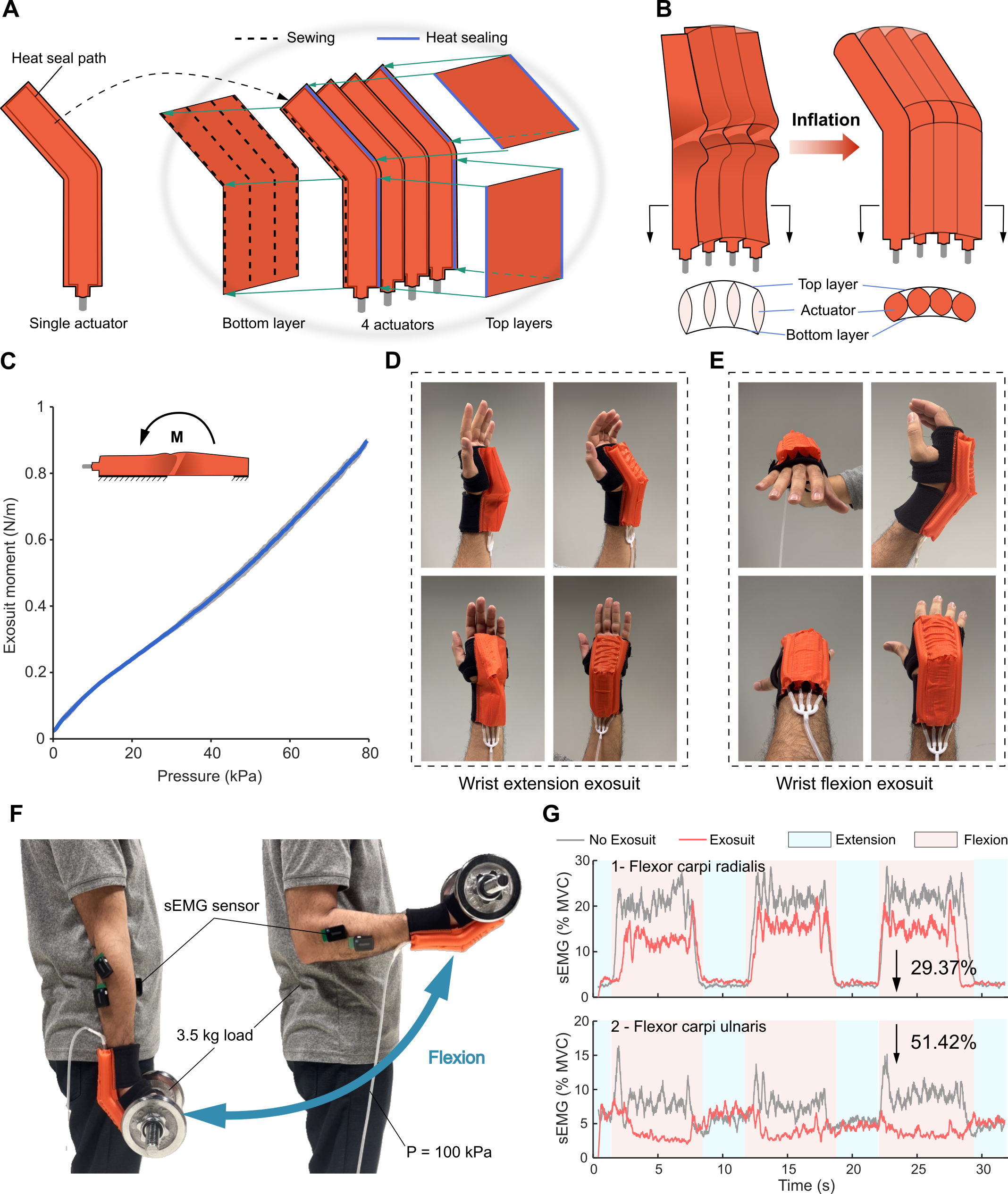}
    \caption{\textbf{Lightweight soft exoskeleton demonstrates therapeutic efficacy.} \textbf{(A)--(B)} Exoskeleton design incorporating parallel GPAs architecture. \textbf{(C)} Linear torque-pressure relationship enables predictable assistance. \textbf{(D)--(E)} Bidirectional functionality for flexion and extension support. \textbf{(F)} Dynamic testing protocol with EMG monitoring. \textbf{(G)} Significant muscle activity reduction validates therapeutic potential.}
    \label{Fig-M-Exosuit}
\end{figure}

We developed a soft wrist exoskeleton (49~g total mass of wearable components) utilizing parallel GPAs architecture to validate therapeutic applications (Fig.~\ref{Fig-M-Exosuit}A--B, Supplementary Fig.~S6 and Movie S3). The exoskeleton integrated parallel GPAs with two fabric pieces featuring Velcro fasteners for secure mounting—one for the hand and another for the forearm. Using the characterization setup shown in Fig.~S6E, we measured output moment across the operational range. The actuator delivered substantial torque assistance (0.9~Nm at 80~kPa) with linear pressure-torque relationships enabling predictable assistance modulation (Fig.~\ref{Fig-M-Exosuit}C). Bidirectional functionality accommodated both flexion and extension assistance through strategic actuator placement modifications (Fig.~\ref{Fig-M-Exosuit}D--E).

We conducted a preliminary user study with a healthy male subject (35~years) following institutional ethical approval to validate exoskeleton assistance efficacy. The subject performed dynamic wrist flexion and isometric testing protocols while supporting a 3.5 kg load in hand (Fig.~\ref{Fig-M-Exosuit}F and Supplementary Fig.~S6). Surface electromyographic (sEMG) sensors monitored muscle activity throughout testing to quantify assistance effects. The exoskeleton achieved substantial muscle activity reductions: 29.37\% for flexor carpi radialis and 51.42\% for flexor carpi ulnaris during dynamic testing (Fig.~\ref{Fig-M-Exosuit}G), with comparable reductions (28.84\% and 28.65\%, respectively) during isometric trials (Supplementary Fig.~S6F and G).

These substantial reductions in muscle activation demonstrate the exoskeleton's efficacy in reducing user effort during loaded wrist movements, validating GPAs potential for rehabilitation and assistance applications. The lightweight, ergonomic, and compact design with predictable assistance patterns addresses critical requirements for practical wearable therapeutic devices\cite{Rui2025_LPPAMs_actuator_app}.

\subsection*{Soft kinesthetic haptic interface for immersive interaction}

\begin{figure}[htbp]
    \centering
    \includegraphics[width=1\textwidth]{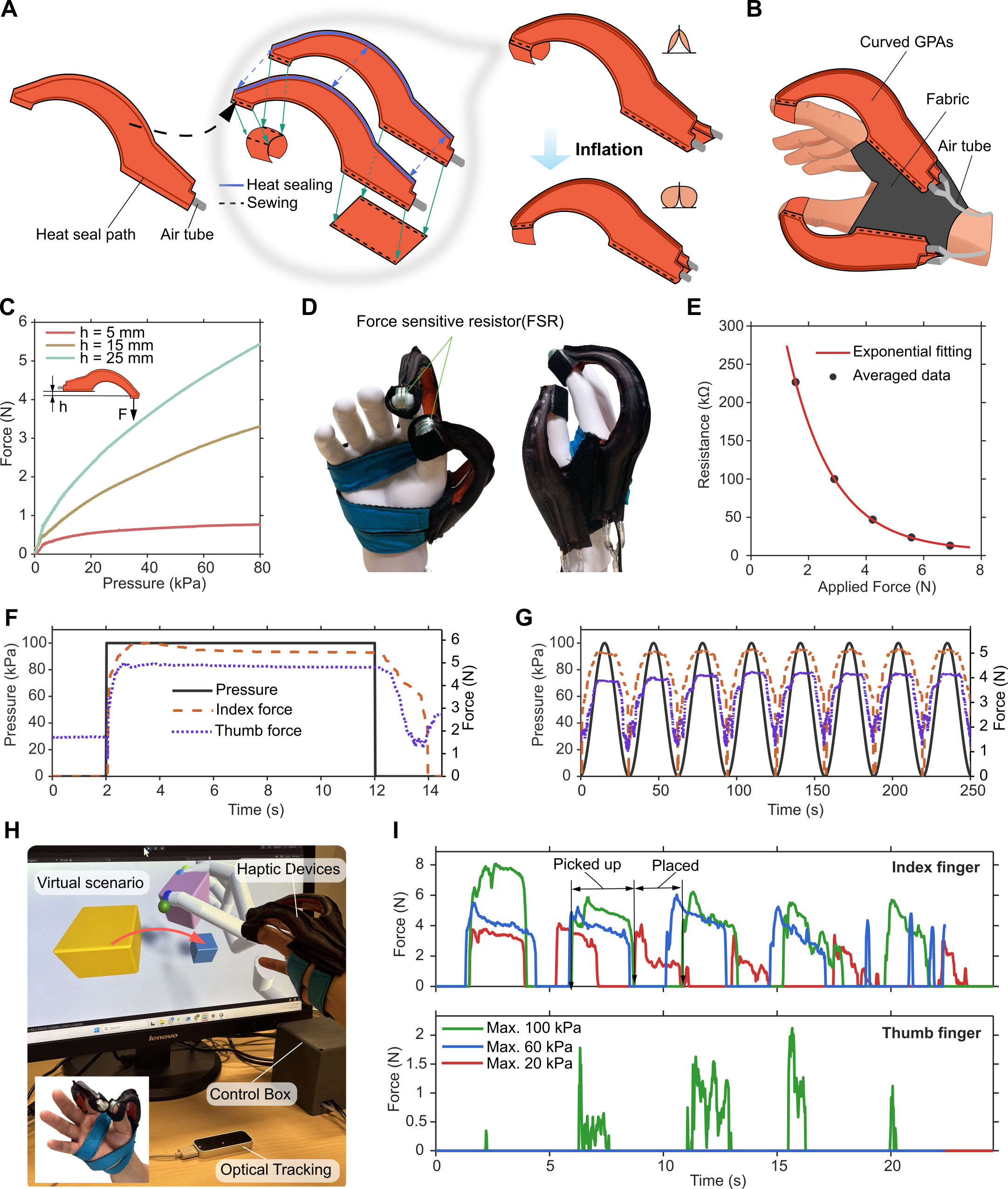}
    \caption{\textbf{Kinesthetic haptic interface enables immersive force feedback.} \textbf{(A)--(B)} Lightweight device architecture with curved GPAs. \textbf{(C)} Force-pressure-displacement relationships for predictable output control. \textbf{(D)--(E)} Dynamic characterization setup with integrated force sensing and fitted force-resistance relationship. \textbf{(F)--(G)} Rapid response characteristics and sinusoidal tracking performance. \textbf{(H)--(I)} Virtual reality integration demonstrating practical manipulation feedback.}
    \label{Fig-M-Haptics}
\end{figure}

We developed a lightweight soft haptic device (30.8~g of wearable components) for thumb and index finger kinesthetic feedback to demonstrate GPAs capabilities for human-computer interaction (Fig.~\ref{Fig-M-Haptics}A--B, Supplementary Fig.~S7, and Movie S4). The haptic device featured two curved GPAs for thumb and index finger, mounted to the hand via fabric with Velcro fasteners (the fabric was dyed black to optimize optical tracking performance). The arched architecture of curved GPAs strategically avoided interference with natural finger motion while delivering targeted kinesthetic feedback at the fingertips.

Force characterization revealed pressure- and displacement-dependent behavior, achieving maximum output forces of 5.4~N at 80~kPa with 25 mm displacement (Fig.~\ref{Fig-M-Haptics}C and Fig.~S8D). Dynamic testing employed a hand mannequin with an integrated force-sensitive resistor (FSR) as the force sensor (Fig.~\ref{Fig-M-Haptics}D). Based on the empirically fitted resistance-force relationship of the FSR (Fig.~\ref{Fig-M-Haptics}E), we estimated interaction forces between finger and haptic device. The haptic device demonstrated rapid step response characteristics: 0.06-s delay time and 0.24 s rise time for both finger actuators (Fig.~\ref{Fig-M-Haptics}F). Dynamic sinusoidal force tracking tests revealed robust performance, with the index finger achieving a 0--5.1~N force range and the thumb achieving a 1.2--4.2~N range (Fig.~\ref{Fig-M-Haptics}G).

Integration of the haptic device with virtual reality environments represents a practical utility for immersive manipulation tasks. We preliminarily tested force feedback capabilities in a virtual pick-and-place task scenario. The subject performed repeated transfers of a purple object from a yellow base to a blue target (Fig.~\ref{Fig-M-Haptics}H). During task execution, the device provided forces reaching up to 8~N for the index finger and up to 8~N for the thumb during object interaction scenarios (Fig.~\ref{Fig-M-Haptics}I). Variable pressure mapping enabled simulation of different interaction forces under identical virtual contact conditions, which can be leveraged for simulating materials with varying stiffness properties.

\subsection*{Multi-gait bipedal locomotion robot}

\begin{figure}[htbp]
    \centering
    \includegraphics[width=1\textwidth]{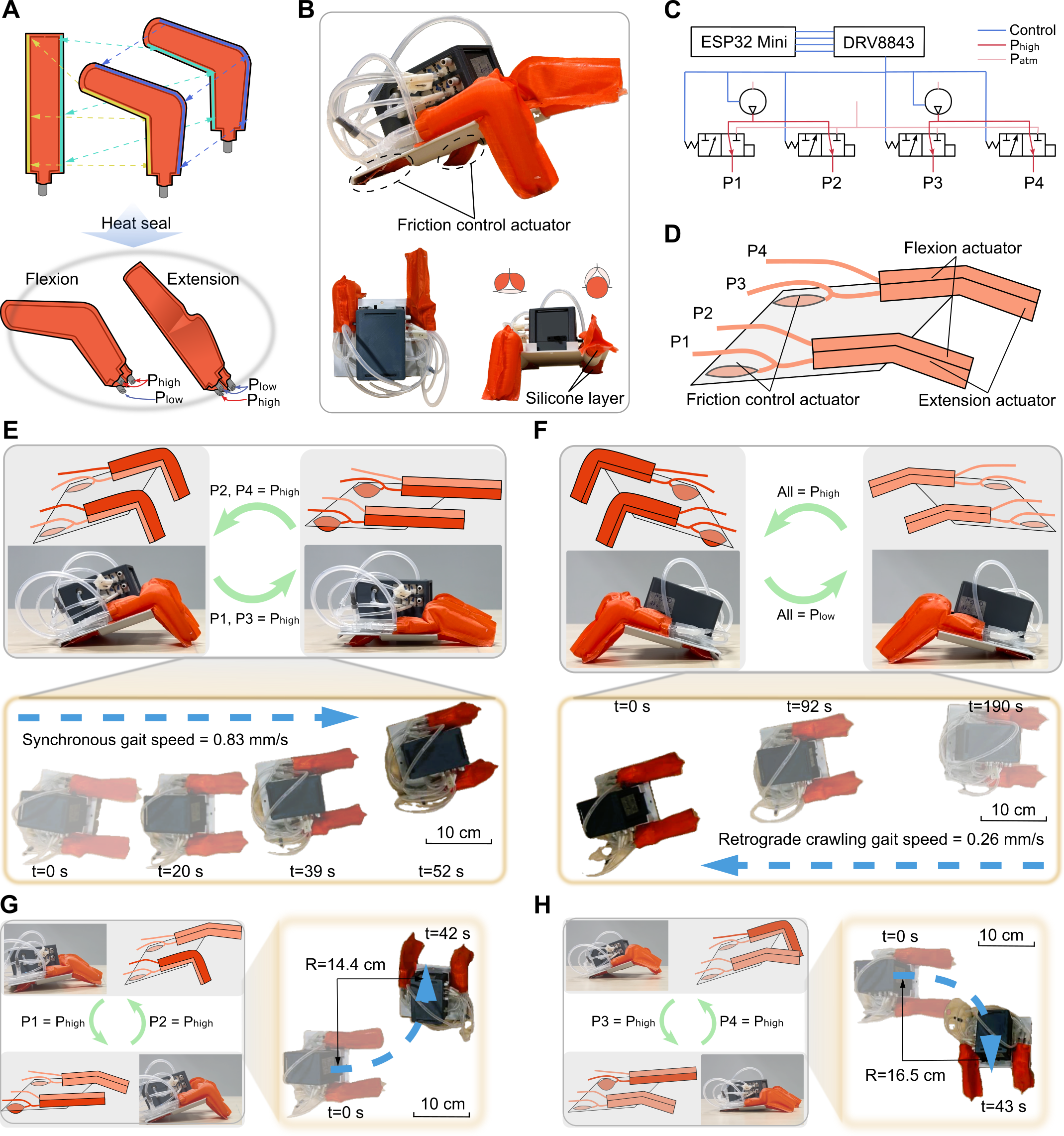}
    \caption{\textbf{Multi-gait bipedal robot demonstrates autonomous locomotion capabilities.} \textbf{(A)--(B)} Dual GPAs leg architecture with integrated friction control. \textbf{(C)--(D)} Pneumatic control logic enabling three-state leg operation. \textbf{(E)--(F)} Synchronous forward gait and retrograde crawling patterns. \textbf{(G)--(H)} Unilateral turning gaits for precise navigation control.}
    \label{Fig-M-robot}
\end{figure}

We constructed a bipedal walking robot (208~g total mass including 164 g control system) featuring dual GPAs legs with integrated friction control chambers to demonstrate autonomous locomotion capabilities (Fig.~\ref{Fig-M-robot}A--B, Supplementary Fig.~S9 and Movie S5). Each leg incorporated flexion and extension chambers, while friction control actuators shared pneumatic actuation with extension chambers for synchronized operation. Four independent pneumatic ports connected to the two legs provided maximum 60 kPa pressure actuation. Each leg operated in three distinct states—flexion, extension, and relaxation—depending on the control system input (Fig.~\ref{Fig-M-robot}C--D).

The friction control system actively modulated ground contact through pressure-dependent surface interaction. Deflated friction chambers minimized ground interaction via reduced contact area, while inflated chambers maximized friction through engagement of silicone-coated surfaces. This mechanism enabled sophisticated gait control without requiring complex mechanical components.

Systematic gait programming produced four distinct locomotion modes: synchronous forward motion, retrograde crawling, and unilateral turning in both directions. Synchronous forward motion achieved the highest speed at 0.83~mm/s (Fig.~\ref{Fig-M-robot}E), while retrograde crawling operated at 0.26~mm/s (Fig.~\ref{Fig-M-robot}F). Unilateral actuation patterns enabled turning maneuvers with small turning radii (14.4~cm left, 16.5~cm right) (Fig.~\ref{Fig-M-robot}G--H), demonstrating the system's maneuverability for autonomous navigation applications.

\section*{Discussion}

We introduce GPAs as a novel design paradigm that addresses critical limitations hindering practical deployment of fabric-based pneumatic systems in soft robotics. Our constraint layer architecture eliminates the deformation instabilities that have limited practical implementation of conventional single-chamber designs, hence enabling unprecedented geometric complexity through programmable CNC heat-sealing and cutting processes. This represents a fundamental advancement beyond incremental improvements, establishing a new foundation for predictable, high-performance soft actuator design.

The key features achieved by GPAs—including stable and predictable deformation behavior, near-zero bending radii with sharp geometric transitions, multi-states actuation capabilities, and customizable geometries—collectively represent a paradigm shift from conventional pneumatic actuators. The capability for minimal bending radii directly addresses spatial constraints inherent to wearable applications, where compact form factors and precise motion control are essential for practical deployment. The flat or concave contact surfaces enabled by strategic constraint layer integration offer particular advantages for ergonomic compatibility compared to the inherently convex profiles of conventional designs. Extended GPAs configurations further demonstrate possibilities for compact multi-states and bidirectional actuation, substantially expanding potential applications across diverse operational contexts.

Our mathematical model of GPAs behavior, validated through systematic characterization experiments, provides a foundation for automated design optimization and performance prediction across diverse application requirements. The model, together with comprehensive bench characterization of prototypes, establishes key GPAs features, particularly the linear relationship between initial angle $\alpha_0$ and final inflated angle $\alpha_1$ (Fig.~\ref{Fig-M-Characterzation}E), and the nonlinear torque-angle relationship (Fig.~\ref{Fig-M-Characterzation}H). These predictable relationships enable rational design of GPAs tailored to specific performance requirements.

Validation across three distinct application domains demonstrated GPAs versatility and performance advantages relative to existing approaches. The 49 g compact soft exoskeleton achieved substantial muscle activity reduction (up to 51\%) while maintaining natural motion patterns—a critical requirement for therapeutic applications\cite{Singn2023Wrist,Sepehri2024Wrist}. 
The 30.8 g haptic interface delivered coherent kinesthetic feedback (up to 8~N) with rapid response characteristics suitable for immersive virtual reality interactions. Compared with other soft kinesthetic feedback devices\cite{connelly2010_pneumatic_haptic_kinethetic_fabric,jadhav2017_haptic_kinethetic_silicone,wang2018_haptic_kinethetic_review}, the proposed curved actuator architecture enables targeted haptic feedback delivery at fingertips without introducing distributed contact forces along the finger dorsum. This configuration more closely resembles realistic and natural contact conditions.
The bipedal robot demonstrated autonomous locomotion capabilities through a structured gait programming, showing GPAs potential for mobile robotic systems. These diverse implementations collectively showcase how the core GPAs features—deformation stability, customizable geometry, minimal bending radii, and multi-states actuation—enable transformative capabilities that extend beyond the limitations of conventional fabric-based pneumatic actuators.


Future development directions include integration of embedded sensing for closed-loop control, exploration of advanced materials for enhanced performance characteristics, and investigation of hybrid architectures combining GPAs with complementary actuation modalities for expanded functional capabilities.





\section*{Materials and Methods}

\subsection*{Materials and fabrication}

Thermoplastic polyurethane (TPU)-coated 40D ripstop nylon fabric (Adventure Expert) served as the primary actuator material throughout this study. For fabrication, we adapted a commercial 3D printer platform (Anycubic Mega) into a CNC heat-sealing machine as shown in Fig.~\ref{Fig-M-Illustration}B. To enable CNC heat sealing of single-chamber actuators with designed shapes, the deflation geometry was first inversely calculated based on the target inflated geometry and the linear angular variation pattern observed during inflation. Subsequently, the geometry was modeled in commercial three-dimensional computer-aided design software and imported into the CNC heat-sealing machine for thermal fabrication. A TPU air tube was connected to the chamber using cyanoacrylate adhesive for pneumatic access. Finally, two pieces of single-chamber actuator were heat sealed (or sewn) together with a constraint layer manually to construct a complete GPA.

\subsection*{Pneumatic control system}

A pneumatic pressure supply utilized a commercial air compressor (FIAC F6000/50) delivering consistent pressure up to 800~kPa with minimal fluctuation. Proportional solenoid valves (ITV0010, SMC) provided precise pressure regulation, while an ESP32 microcontroller (Espressif Systems) enabled real-time control with integrated pressure sensor feedback. Dynamic pressure modulation employed 3~kPa/s rates for consistent experimental protocols across all testing conditions.

\subsection*{Performance characterization and modeling}

Before comprehensive modeling and characterization, we experimentally demonstrated that the GPAs design transforms unpredictable single-chamber behavior into stable, repeatable actuation patterns. We compared a GPAs ($L_1 = 80$~mm, $L_0 = 24$~mm, $\alpha_0 = 120°$, and $W = 20$~mm) with a single-chamber actuator ($L_1 = 80$~mm, $L_0 = 24$~mm, $\beta_0 = 120°$) under external deformation as depicted in Fig.~\ref{Fig-M-Illustration}C. These actuators were fixed to the end of a robotic arm and moved downward against a horizontal plane at 3~mm/s. The single-chamber actuator underwent substantial lateral deviation, while the GPAs exhibited stable behavior without deviations outside the actuation plane. This behavior was repeatable across multiple independent trials ($n > 10$).

Angular measurements utilized calibrated digital protractors for precise geometric analysis; the average result of three repetitions is presented as the final result for each condition. To investigate parameter effects systematically, we prepared actuators with different circumferential lengths (19~mm, 24~mm, 29~mm, and 34~mm), initial angles $\beta_0$ (60°, 90°, 120°, and 150°), and actuator widths (15~mm, 20~mm, 25~mm, 30~mm), while default parameters were 19~mm for circumferential length, 90° for initial angle, and 20~mm for actuator width. All experimental data and design files are available in supplementary materials, where all .stl files for CNC fabrication are provided.

Torque characterization employed a servo motor (CubeMars AK80-8) for precise angular control, while force measurements utilized calibrated LSB205 sensors (Futek) at 100~Hz sampling frequency. The motor rotated starting from a preset angle (depending on the initial angle, where the actuator had no contact with the force sensor when inflated) at 3°/s rotation rate until the angle decreased to 180° and then returned to the preset position. Data processing included 20-point moving window filtration for noise reduction and artifact removal.

We developed a mathematical model to predict behavior of the actuator based on assumptions of the folding process when subjected to external loads. The model is detailed in supplementary text\cite{o2022unfolding_math}. Additional comparisons between modeling predictions and experimental results with different parameter combinations are shown in Fig.~S5.

\subsection*{Experiments for the wrist exoskeleton}

The parallel actuator consisted of four single-chamber actuators ($L_1 = 83$~mm, $L_0 = 27$~mm, $\beta_0 = 140°$) with constraint layers at the bottom and top, while one bottom layer and two top layers shared the same actuator width $w = 50$~mm. The actuator was then sewn to two fabric pieces with Velcro fasteners—one for the hand and another for the forearm—for torque transmission (Fig.~S6A). Using the setup shown in Fig.~\ref{Fig-M-Exosuit}C and Fig.~S6E, the actuator was mounted at 180° bending angle, and pressure was then increased slowly to 80~kPa while reaction force was recorded continuously.

Exoskeleton validation employed a healthy male subject (35~years) with written informed consent under ethical approval from Scuola Superiore Sant'Anna institutional review board (approval number 40/2025). Testing protocols included repetitive flexion tasks (3.5 kg load, 5 s hold periods) and isometric endurance trials (1 minute duration) with comprehensive EMG monitoring using Delsys Trigno wireless systems. Muscle activity measurements targeted wrist flexor and extensor muscles (Flexor Carpi Radialis, Flexor Carpi Ulnaris, Extensor Carpi Radialis Longus, Extensor Carpi Radialis Brevis) with 1259~Hz sampling frequency and maximum voluntary contraction (MVC) normalization. Before formal experiments, MVC was recorded for muscle activity normalization, and average MVC during flexion was used to quantify assistance-induced reduction.

\subsection*{Haptic interface characterization}

The haptic device consisted of two curved actuators, a small constraint layer at the tip functioning as a finger cap, and a larger constraint structure for secure hand mounting (Fig.~S7A). The curved geometry strategically avoided unwanted forces on the dorsal side of the finger, especially at joint locations, enabling more natural hand grasping and pinch force feedback for realistic haptic experience (Fig.~S7C and D).

Using the setup shown in Fig.~\ref{Fig-M-Haptics}C and Fig.~S8D, fingertip force was estimated through resistance measurements. Considering the effect of flexion angle on force output, we varied the height clearance between the fingertip and hand baseline. Force sensing utilized flexible resistance sensors (FSR 104-MF01-N-220-A02, Taiwan Alpha Electronic) with voltage division circuits for resistance measurement. To determine the force-resistance relationship of the FSR, we designed a calibration setup with a real human finger (Fig.~S8A) and measured FSR resistance at different applied forces for three repetitions. The averaged resistance under given force was used for fitting (Fig.~S8B and C), and exponential fitting was selected based on comparison results of different fitting methods (Fig.~\ref{Fig-M-Haptics}E). With fitted force-resistance relationships established across operational ranges, we estimated interaction force at the fingertip first with a hand mannequin (Fig.~\ref{Fig-M-Haptics}D), including step response testing and sinusoidal force tracking testing (Fig.~\ref{Fig-M-Haptics}F and G).

One healthy subject participated in a preliminary virtual reality manipulation task with written informed consent under ethical approval from Scuola Superiore Sant'Anna institutional review board (protocol 412023). In the task, we utilized a virtual scenario similar to our previous work (Fig.~S8G)\cite{Rui2025ThermalHaptic}. A portable pneumatic control box with two independent pneumatic channels was used to modulate pressure, as shown in Fig.~S8E. To reflect hand gestures into virtual reality in real time, we used a commercial optical hand tracking system (Leap Motion, Ultraleap) and dyed the orange fabric black for improved optical tracking performance. During operation, reaction force between the virtual hand and object was estimated using virtual balls positioned on fingertips\cite{leonardis2016_Haptic_Rendering}, and estimated force was then transmitted to the control box using UDP protocol. The control box mapped force from a minimum 1~kPa baseline to maximum mapping pressure using proportional valves. The minimum 1~kPa pressure maintained the actuator in an inflated state and reduced subsequent inflation time, while maximum mapping pressure determined the simulated stiffness characteristic of the haptic device.

In experiments, subjects were asked to transfer a purple object from a yellow base to a blue target for multiple repetitions under different maximum mapping pressures (20~kPa, 60~kPa, and 100~kPa). FSR resistance between the haptic device and fingertip was recorded continuously during the task, and reaction force was estimated and presented based on the force-resistance relationship established previously.

\subsection*{Bipedal robot control}

The bipedal robot consisted of two legs, a 4-channel pneumatic control box, and a 3D-printed mounting layer to secure the actuator and control box. Each actuator leg had two primary states: flexion and extension, determined by selective chamber inflation, and each leg tip incorporated a silicone layer to increase ground friction (Fig.~\ref{Fig-M-robot}A and B). To actively modulate friction during movements, we introduced a friction control actuator with an additional silicone layer at the robot's distal end. The friction control actuator was pneumatically connected to the extension chamber so that it exhibited higher friction when the extension chamber was inflated, enabling synchronized friction modulation.

The control box integrated two mini pumps (KPM14A, Koge), four mini valves (S070, SMC), and a control board with an ESP32 microcontroller and three DRV8833 motor control modules. With four independently controlled valves, the actuator system provided four pneumatic output ports, while output pressure could be modulated by controlling pulse-width modulation (PWM) of pumps (40 kPa to 60 kPa). By systematically varying actuator inflation patterns, different gaits were achieved for diverse locomotion behaviors.


\clearpage 

%
\bibliography{mybib} 

\begin{thebibliography}{10}
\providecommand{\url}[1]{\texttt{#1}}
\expandafter\ifx\csname urlstyle\endcsname\relax
  \providecommand{\doi}[1]{doi:\discretionary{}{}{}#1}\else
  \providecommand{\doi}{doi:\discretionary{}{}{}\begingroup \urlstyle{rm}\Url}\fi

\bibitem{el2020_Soft_Actuator_Review}
N.~El-Atab, \emph{et~al.}, Soft actuators for soft robotic applications: A review. \emph{Advanced Intelligent Systems} \textbf{2}~(10), 2000128 (2020).

\bibitem{zhou2024_portable_exoskeleton_app}
Y.~M. Zhou, \emph{et~al.}, A portable inflatable soft wearable robot to assist the shoulder during industrial work. \emph{Science Robotics} \textbf{9}~(91), eadi2377 (2024).

\bibitem{Rui2025_Glove_app}
R.~Chen, D.~Leonardis, A.~Frisoli, D.~Chiaradia, A Powerful Customized Fabric-Based Soft Robotic Glove for Assistance and Rehabilitation, in \emph{2025 International Conference On Rehabilitation Robotics (ICORR)} (2025), pp. 669--674, \doi{10.1109/ICORR66766.2025.11063169}.

\bibitem{ze2022S_pinning_Origami_locomotion_medical_app}
Q.~Ze, \emph{et~al.}, Spinning-enabled wireless amphibious origami millirobot. \emph{Nature communications} \textbf{13}~(1), 3118 (2022).

\bibitem{choi2022_Body_Integreted_medical_app}
Y.~S. Choi, \emph{et~al.}, A transient, closed-loop network of wireless, body-integrated devices for autonomous electrotherapy. \emph{Science} \textbf{376}~(6596), 1006--1012 (2022).

\bibitem{guo2024_DEA_haptic_APP}
Y.~Guo, \emph{et~al.}, Haptic artificial muscle skin for extended reality. \emph{Science Advances} \textbf{10}~(43), eadr1765 (2024).

\bibitem{du2024_haptiknit_Haptic_app}
C.~du~Pasquier, \emph{et~al.}, Haptiknit: Distributed stiffness knitting for wearable haptics. \emph{Science Robotics} \textbf{9}~(97), eado3887 (2024).

\bibitem{ha2025_full_freedom_Haptic_app}
K.-H. Ha, \emph{et~al.}, Full freedom-of-motion actuators as advanced haptic interfaces. \emph{Science} \textbf{387}~(6741), 1383--1390 (2025).

\bibitem{zhang_2023_Origami_Active_Haptic_app}
Z.~Zhang, \emph{et~al.}, Active mechanical haptics with high-fidelity perceptions for immersive virtual reality. \emph{Nature Machine Intelligence} \textbf{5}~(6), 643--655 (2023).

\bibitem{katzschmann2018_hydrolic_Underwater_sea_app}
R.~K. Katzschmann, J.~DelPreto, R.~MacCurdy, D.~Rus, Exploration of underwater life with an acoustically controlled soft robotic fish. \emph{Science Robotics} \textbf{3}~(16), eaar3449 (2018).

\bibitem{pan2025_locomotion_sea_app}
F.~Pan, \emph{et~al.}, Miniature deep-sea morphable robot with multimodal locomotion. \emph{Science Robotics} \textbf{10}~(100), eadp7821 (2025).

\bibitem{zhakypov2019_Multi_Locomotion_locomotion_app}
Z.~Zhakypov, K.~Mori, K.~Hosoda, J.~Paik, Designing minimal and scalable insect-inspired multi-locomotion millirobots. \emph{Nature} \textbf{571}~(7765), 381--386 (2019).

\bibitem{shepherd2011_pneumatic_multigait_Locomotion_app}
R.~F. Shepherd, \emph{et~al.}, Multigait soft robot. \emph{Proceedings of the national academy of sciences} \textbf{108}~(51), 20400--20403 (2011).

\bibitem{laschi2012_Soft_Octopus_app}
C.~Laschi, \emph{et~al.}, Soft robot arm inspired by the octopus. \emph{Advanced robotics} \textbf{26}~(7), 709--727 (2012).

\bibitem{fu2025_origami_manipulation_app}
K.~Fu, \emph{et~al.}, Origami exoskeletons for enhanced soft robotic manipulation. \emph{Science Advances} \textbf{11}~(31), eadv6629 (2025).

\bibitem{wu2021_sStretchable_Origami_arm_app}
S.~Wu, \emph{et~al.}, Stretchable origami robotic arm with omnidirectional bending and twisting. \emph{Proceedings of the National Academy of Sciences} \textbf{118}~(36), e2110023118 (2021).

\bibitem{youn2025_DEA_Haptic_app}
J.-H. Youn, \emph{et~al.}, Skin-attached haptic patch for versatile and augmented tactile interaction. \emph{Science Advances} \textbf{11}~(12), eadt4839 (2025).

\bibitem{aubin2023_combustion_actuators}
C.~A. Aubin, \emph{et~al.}, Powerful, soft combustion actuators for insect-scale robots. \emph{Science} \textbf{381}~(6663), 1212--1217 (2023).

\bibitem{fonseca2025Electrically_phase_transition_actuators}
D.~Fonseca, P.~Neto, Electrically-driven phase transition actuators to power soft robot designs. \emph{Nature Communications} \textbf{16}~(1), 3920 (2025).

\bibitem{mao2020_Soft_Electromagnetic_actuator}
G.~Mao, \emph{et~al.}, Soft electromagnetic actuators. \emph{Science advances} \textbf{6}~(26), eabc0251 (2020).

\bibitem{zhang2025D_Printed_Magnetoactive_Origami_Actuator}
S.~Zhang, \emph{et~al.}, 3D-Printed Soft Magnetoactive Origami Actuators. \emph{Advanced Functional Materials} p. e16404 (2025).

\bibitem{yang2025_liquid_Crystal_Elastomer_Fiber_Actuators}
H.~Yang, \emph{et~al.}, Weaving liquid crystal elastomer fiber actuators for multifunctional soft robotics. \emph{Science Advances} \textbf{11}~(8), eads3058 (2025).

\bibitem{li2021_Liquid_Crystal_Elastomers_muscles_actuator}
S.~Li, \emph{et~al.}, Digital light processing of liquid crystal elastomers for self-sensing artificial muscles. \emph{Science Advances} \textbf{7}~(30), eabg3677 (2021).

\bibitem{gravert2024_electrohydraulic_actuators}
S.-D. Gravert, \emph{et~al.}, Low-voltage electrohydraulic actuators for untethered robotics. \emph{Science Advances} \textbf{10}~(1), eadi9319 (2024).

\bibitem{Rui2025_LPPAMs_actuator_app}
R.~Chen, A.~Frisoli, D.~Chiaradia, Lateral Pleated Pneumatic Artificial Muscles: High-Contraction and Extended High-Force Range Actuators for Soft Robotics. \emph{IEEE/ASME Transactions on Mechatronics} pp. 1--11 (2025), \doi{10.1109/TMECH.2025.3593667}.

\bibitem{tang2020_pneumatic_bistable_Spine_locomotion_app}
Y.~Tang, \emph{et~al.}, Leveraging elastic instabilities for amplified performance: Spine-inspired high-speed and high-force soft robots. \emph{Science advances} \textbf{6}~(19), eaaz6912 (2020).

\bibitem{feng2023_X-PAMs_actuator}
M.~Feng, D.~Yang, L.~Ren, G.~Wei, G.~Gu, X-crossing pneumatic artificial muscles. \emph{Science advances} \textbf{9}~(38), eadi7133 (2023).

\bibitem{niiyama2015_Pouch_Motor}
R.~Niiyama, \emph{et~al.}, Pouch motors: Printable soft actuators integrated with computational design. \emph{Soft Robotics} \textbf{2}~(2), 59--70 (2015).

\bibitem{xavier2022Pneuamtic_review}
M.~S. Xavier, \emph{et~al.}, Soft pneumatic actuators: A review of design, fabrication, modeling, sensing, control and applications  (2022).

\bibitem{zhu2020_Sewing_fabrication_tube}
M.~Zhu, T.~N. Do, E.~Hawkes, Y.~Visell, Fluidic fabric muscle sheets for wearable and soft robotics. \emph{Soft robotics} \textbf{7}~(2), 179--197 (2020).

\bibitem{phan2022Fabrication_sewing_textile}
P.~T. Phan, \emph{et~al.}, Smart textiles using fluid-driven artificial muscle fibers. \emph{Scientific reports} \textbf{12}~(1), 11067 (2022).

\bibitem{guo2024_Encoded_Sewing_actuator}
X.~Guo, \emph{et~al.}, Encoded sewing soft textile robots. \emph{Science Advances} \textbf{10}~(1), eadk3855 (2024).

\bibitem{luo2022Fabrication_Machine_knitting}
Y.~Luo, \emph{et~al.}, Digital fabrication of pneumatic actuators with integrated sensing by machine knitting, in \emph{Proceedings of the 2022 CHI Conference on Human Factors in Computing Systems} (2022), pp. 1--13.

\bibitem{sanchez2023_3D_knitting}
V.~Sanchez, \emph{et~al.}, 3D knitting for pneumatic soft robotics. \emph{Advanced functional materials} \textbf{33}~(26), 2212541 (2023).

\bibitem{cappello2018FabricationAnisotropy}
L.~Cappello, \emph{et~al.}, Exploiting textile mechanical anisotropy for fabric-based pneumatic actuators. \emph{Soft robotics} \textbf{5}~(5), 662--674 (2018).

\bibitem{Li_2024_knitting}
H.~Li, \emph{et~al.}, Yarn-grouping weaving soft robotics with directional inflation, bilateral bending, and self-sensing for healthcare. \emph{Cell Reports Physical Science} \textbf{5}~(8), 102137 (2024), \doi{https://doi.org/10.1016/j.xcrp.2024.102137}, \url{https://www.sciencedirect.com/science/article/pii/S2666386424004211}.

\bibitem{amiri2018laser_cut}
A.~A. Amiri~Moghadam, \emph{et~al.}, Laser cutting as a rapid method for fabricating thin soft pneumatic actuators and robots. \emph{Soft robotics} \textbf{5}~(4), 443--451 (2018).

\bibitem{yamaoka2018_accordionfab_Laser_cut}
J.~Yamaoka, \emph{et~al.}, AccordionFab: Fabricating inflatable 3D objects by laser cutting and welding multi-layered sheets, in \emph{Adjunct Proceedings of the 31st Annual ACM Symposium on User Interface Software and Technology} (2018), pp. 160--162.

\bibitem{Wu2025pouch_stack}
J.~Wu, M.~Wu, C.~Wang, G.~Xie, Monolithic Programmable Fabric-Stacking Enables Multifunctional Soft Robots. \emph{IEEE Transactions on Robotics} \textbf{41}, 4810--4828 (2025), \doi{10.1109/TRO.2025.3593118}.

\bibitem{zhang2023_Soft_Fabric_actuator}
Z.~Zhang, \emph{et~al.}, Soft and lightweight fabric enables powerful and high-range pneumatic actuation. \emph{Science Advances} \textbf{9}~(15), eadg1203 (2023).

\bibitem{nishioka2017Fabric_pleated}
Y.~Nishioka, \emph{et~al.}, Development of a pneumatic soft actuator with pleated inflatable structures. \emph{Advanced Robotics} \textbf{31}~(14), 753--762 (2017).

\bibitem{suulker2022soft_pouch_motor}
C.~Suulker, S.~Skach, K.~Althoefer, Soft robotic fabric actuator with elastic bands for high force and bending performance in hand exoskeletons. \emph{IEEE Robotics and Automation Letters} \textbf{7}~(4), 10621--10627 (2022).

\bibitem{rajappan2022_logic_pouch_motor}
A.~Rajappan, \emph{et~al.}, Logic-enabled textiles. \emph{Proceedings of the National Academy of Sciences} \textbf{119}~(35), e2202118119 (2022).

\bibitem{o2022unfolding_cylinder}
C.~T. O'Neill, C.~M. McCann, C.~J. Hohimer, K.~Bertoldi, C.~J. Walsh, Unfolding textile-based pneumatic actuators for wearable applications. \emph{Soft Robotics} \textbf{9}~(1), 163--172 (2022).

\bibitem{yang2020_design_chamber_cylinder}
H.~D. Yang, A.~T. Asbeck, Design and characterization of a modular hybrid continuum robotic manipulator. \emph{IEEE/ASME Transactions on Mechatronics} \textbf{25}~(6), 2812--2823 (2020).

\bibitem{gohlke2023wireshape_CNC}
K.~Gohlke, H.~Waldsch{\"u}tz, E.~Hornecker, WireShape-A Hybrid Prototyping Process for Fast \& Reliable Manufacturing of Inflatable Interface Props with CNC-Fabricated Heat--Sealing Tools, in \emph{Proceedings of the 8th ACM Symposium on Computational Fabrication} (2023), pp. 1--8.

\bibitem{goshtasbi2025weld_CNC}
A.~Goshtasbi, \emph{et~al.}, Weld n’Cut: Automated fabrication of inflatable fabric actuators, in \emph{2025 IEEE 8th International Conference on Soft Robotics (RoboSoft)} (IEEE) (2025), pp. 1--6.

\bibitem{rus2015_Design_Fabrication_review}
D.~Rus, M.~T. Tolley, Design, fabrication and control of soft robots. \emph{Nature} \textbf{521}~(7553), 467--475 (2015).

\bibitem{Andrade2025_Star_griper}
I.~Andrade-Silva, J.~Marthelot, Fabric-Based Star Soft Robotic Gripper. \emph{Advanced Intelligent Systems} \textbf{5}~(8), 2200435 (2023), \doi{https://doi.org/10.1002/aisy.202200435}, \url{https://advanced.onlinelibrary.wiley.com/doi/abs/10.1002/aisy.202200435}.

\bibitem{Singn2023Wrist}
I.~Singh, \emph{et~al.}, Development of Soft Pneumatic Actuator Based Wrist Exoskeleton for Assistive Motion, in \emph{2023 IEEE/ASME International Conference on Advanced Intelligent Mechatronics (AIM)} (2023), pp. 359--366, \doi{10.1109/AIM46323.2023.10196235}.

\bibitem{Sepehri2024Wrist}
A.~Sepehri, S.~Ward, M.~T. Tolley, T.~K. Morimoto, A Soft Robotic Wrist Orthosis Using Textile Pneumatic Actuators For Passive Rehabilitation, in \emph{2024 IEEE 7th International Conference on Soft Robotics (RoboSoft)} (2024), pp. 284--290, \doi{10.1109/RoboSoft60065.2024.10521992}.

\bibitem{connelly2010_pneumatic_haptic_kinethetic_fabric}
L.~Connelly, \emph{et~al.}, A pneumatic glove and immersive virtual reality environment for hand rehabilitative training after stroke. \emph{IEEE Transactions on Neural Systems and Rehabilitation Engineering} \textbf{18}~(5), 551--559 (2010).

\bibitem{jadhav2017_haptic_kinethetic_silicone}
S.~Jadhav, V.~Kannanda, B.~Kang, M.~T. Tolley, J.~P. Schulze, Soft robotic glove for kinesthetic haptic feedback in virtual reality environments. \emph{electronic imaging} \textbf{29}, 19--24 (2017).

\bibitem{wang2018_haptic_kinethetic_review}
D.~Wang, \emph{et~al.}, Toward whole-hand kinesthetic feedback: A survey of force feedback gloves. \emph{IEEE transactions on haptics} \textbf{12}~(2), 189--204 (2018).

\bibitem{o2022unfolding_math}
C.~T. O'Neill, C.~M. McCann, C.~J. Hohimer, K.~Bertoldi, C.~J. Walsh, Unfolding textile-based pneumatic actuators for wearable applications. \emph{Soft Robotics} \textbf{9}~(1), 163--172 (2022).

\bibitem{Rui2025ThermalHaptic}
R.~Chen, D.~Chiaradia, A.~Frisoli, D.~Leonardis, A Soft Fabric-Based Thermal Haptic Device for VR and Teleoperation (2025), \url{https://arxiv.org/abs/2508.20831}.

\bibitem{leonardis2016_Haptic_Rendering}
D.~Leonardis, M.~Solazzi, I.~Bortone, A.~Frisoli, A 3-RSR haptic wearable device for rendering fingertip contact forces. \emph{IEEE transactions on haptics} \textbf{10}~(3), 305--316 (2016).

\end{thebibliography}
\bibliographystyle{sciencemag}

%
%
%
%
%
%


\section*{Acknowledgments}
We thank the user study participants for their dedicated involvement and colleagues in soft robotics and haptic interfaces for valuable scientific discussions that enhanced this research.

\paragraph*{Funding:}
This research was supported by MSCA-DN Project 101073374 – ReWIRE.

\paragraph*{Author contributions:}
Conceptualization: RC, DL, DC, AF; Methodology: RC, DL; Investigation: RC; Visualization: RC; Writing—original draft: RC; Writing—review \& editing: RC, DL, DC, AF; Funding acquisition: AF; Project administration: DC, AF; Supervision: DC, DL, AF

\paragraph*{Competing interests:}
The authors declare no competing interests.

\paragraph*{Data and materials availability:}
All data supporting the conclusions are available in the main text and supplementary materials. Fabrication protocols, control software, and experimental datasets are available from the corresponding author upon reasonable request.


\subsection*{Supplementary materials}
Supplementary Text\\
Figs. S1 to S9\\
Movies S1 to S5


\newpage


\renewcommand{\thefigure}{S\arabic{figure}}
\renewcommand{\thetable}{S\arabic{table}}
\renewcommand{\theequation}{S\arabic{equation}}
\renewcommand{\thepage}{S\arabic{page}}
\setcounter{figure}{0}
\setcounter{table}{0}
\setcounter{equation}{0}
\setcounter{page}{1} 

\end{document}